%% file: main.tex
\pdfoutput=1

\documentclass{article}

\usepackage[main, final]{neurips_2026}

\usepackage[utf8]{inputenc}
\usepackage[T1]{fontenc}
\usepackage{hyperref}
\usepackage{url}
\usepackage{booktabs}
\usepackage{amsfonts}
\usepackage{nicefrac}
\usepackage{microtype}
\usepackage{xcolor}

\usepackage{graphicx}
\usepackage{subcaption}
\usepackage{multirow}
\usepackage{amsmath}
\usepackage{amssymb}
\usepackage{mathtools}
\usepackage{amsthm}
\usepackage{algorithm}
\usepackage{algorithmic}
\usepackage[capitalize,noabbrev]{cleveref}

\usepackage[disable,textsize=tiny]{todonotes}

\theoremstyle{plain}

\theoremstyle{definition}

\theoremstyle{remark}

\input{math_commands}

\title{Block Sparse Flash Attention}

\author{%
  Daniel Ohayon \\
  Corma \\
  Technion -- Haifa, Israel \\
  {\small\texttt{daniel@corma.ai}} \\
  \And
  Itay Lamprecht\thanks{Work done while at Habana Labs (Intel).} \\
  Crusoe AI \\
  Technion -- Haifa, Israel \\
  {\small\texttt{itayl@campus.technion.ac.il}} \\
  \And
  Itay Hubara \\
  Technion -- Haifa, Israel \\
  {\small\texttt{itayhubara@gmail.com}} \\
  \AND
  Israel Cohen \\
  Technion -- Haifa, Israel \\
  {\small\texttt{icohen@ee.technion.ac.il}} \\
  \And
  Daniel Soudry \\
  Technion -- Haifa, Israel \\
  {\small\texttt{daniel.soudry@gmail.com}} \\
  \And
  Noam Elata \\
  Technion -- Haifa, Israel \\
  {\small\texttt{noamelata@technion.ac.il}} \\
}

\begin{document}

\maketitle

\begin{abstract}
Modern large language models increasingly require long contexts for reasoning and multi-document tasks, but attention's quadratic complexity creates a severe computational bottleneck. We present \emph{Block Sparse Flash Attention} (BSFA), a drop-in replacement that accelerates long-context inference while preserving model quality. Unlike methods that predict importance before computing scores, BSFA computes exact query-key similarities to select the top-$k$ most important value blocks for each query. By comparing per-block maximum scores against calibrated thresholds, we skip approximately 50\% of the computation and memory transfers for pruned blocks. Our training-free approach requires only a one-time threshold calibration on a small dataset to learn the per-layer and per-head attention score distributions. We provide a CUDA kernel implementation that can be used as a drop-in replacement for FlashAttention. On Llama-3.1-8B, BSFA achieves up to $1.13\times$ end-to-end speedup on LongBench with only a $1.1\%$ accuracy drop, and up to $1.24\times$ on Needle-in-a-Haystack retrieval at a $1\%$ accuracy drop. The attention kernel itself accelerates by up to $1.38\times$. We compare BSFA against five recent sparse attention baselines (SpargeAttention, MInference, FlexPrefill, XAttention, and BLASST), and verify the method on Qwen2.5-7B and on A6000 and H100 GPUs. The implementation is available at \url{https://github.com/Danielohayon/Block-Sparse-Flash-Attention}.
\end{abstract}


\section{Introduction}

Large language models have become the backbone of modern AI systems~\citep{brown2020languagemodelsfewshotlearners, touvron2023llamaopenefficientfoundation, achiam2023gpt4technicalreport}, enabling state-of-the-art performance in generation~\citep{achiam2023gpt4technicalreport}, reasoning~\citep{wei2022chainofthoughtpromptingelicitsreasoning, kojima2022largelanguagemodelszeroshot}, and retrieval-augmented applications~\citep{lewis2020retrieval, borgeaud2022improvinglanguagemodelstrieving}. At their core lies scaled dot-product attention~\citep{vaswani2023attentionneed}, whose quadratic compute and storage in the sequence length becomes the dominant bottleneck as practitioners push contexts to hundreds of thousands of tokens, where attention operations consume the majority of inference time and can account for over 90\% of total FLOPs in long-context scenarios. FlashAttention~\citep{dao2022flashattentionfastmemoryefficientexact,dao2023flashattention2fasterattentionbetter} eliminates the memory side of this bottleneck by tiling computation into streaming updates, but the arithmetic work of multiplying every query against every key, applying softmax, then aggregating values remains $O(N^2)$ FLOPs, a cost that in general cannot be made subquadratic~\citep{alman2023fastattentionrequiresbounded,alman2024fundamentallimitationssubquadratic}.

Empirical evidence shows that attention distributions are highly sparse~(Section~\ref{sec:attention-sparsity-patterns}), motivating sparse attention methods that reduce computation by avoiding certain query-key interactions\nopagebreak~\citep{beltagy2020longformerlongdocumenttransformer, roy2020efficientcontentbasedsparseattention, kitaev2020reformerefficienttransformer}. The dominant strategy in this literature is to predict importance \emph{before} observing the scores (via fixed patterns, hashing, clustering, or pooled approximations of $\mathbf{Q}\mathbf{K}^\top$) and commit to pruning decisions on that prediction. Such predictive early pruning risks accuracy degradation when relevant tokens are scattered unpredictably across long contexts.

We propose Block Sparse Flash Attention (BSFA), which inverts this strategy. BSFA computes all query-key scores exactly within FlashAttention's tiled framework, then uses the per-block maximum scores to decide which value blocks to process. Blocks whose maximum scores fall below a calibrated threshold are skipped, so their values are not loaded from HBM, the $\mathbf{P}\mathbf{V}$ multiplication is not performed, and the running softmax statistics are not updated. This exploits the observation that blocks with uniformly low scores contribute negligibly after softmax normalization. If no query-key pair in a block has high enough similarity, the entire block can be skipped without meaningful impact on the output. By preserving exact score computation while selectively skipping the value-side work, BSFA achieves substantial speedups without the accuracy risks of early pruning.

Across our evaluation, BSFA is the only method that reaches the $\geq$99\%-accuracy band on LongBench while accelerating long-context inference and avoiding the slowdowns at short and medium context lengths that affect MInference, FlexPrefill, and XAttention. Experiments on Llama-3.1-8B demonstrate that BSFA achieves up to $1.13\times$ end-to-end speedup on real-world LongBench tasks with only a $1.1\%$ drop from baseline accuracy, and up to $1.24\times$ on Needle-in-a-Haystack retrieval at $99\%$ baseline accuracy. The underlying attention kernel itself is sped up by up to $1.38\times$ at higher sparsity levels (Section~\ref{sec:experiments}). Comparing against five recent sparse attention methods (SpargeAttention~\citep{zhang2025spargeattentionaccuratetrainingfreesparse}, MInference~\citep{jiang2024minference}, FlexPrefill~\citep{lai2025flexprefill}, XAttention~\citep{xu2025xattention}, and BLASST~\citep{yuan2025blasstdynamicblockedattention}), BSFA reaches the $\geq$99\%-accuracy band on LongBench at speedups comparable to MInference and XAttention, while SpargeAttention and FlexPrefill cannot reach this band and BLASST loses more accuracy at matched sparsity. The pre-attention selection methods (MInference, FlexPrefill, XAttention) additionally regress to as low as $0.17\times$ on contexts under 16K, where BSFA stays within a few percent of dense. We additionally verify these results on Qwen2.5-7B and on A6000 and H100 GPUs. We release a CUDA kernel extending FlashAttention-2 as a drop-in replacement.

\begin{figure}[t]
\centering
\begin{subfigure}[t]{0.45\linewidth}
  \centering
  \includegraphics[width=\linewidth]{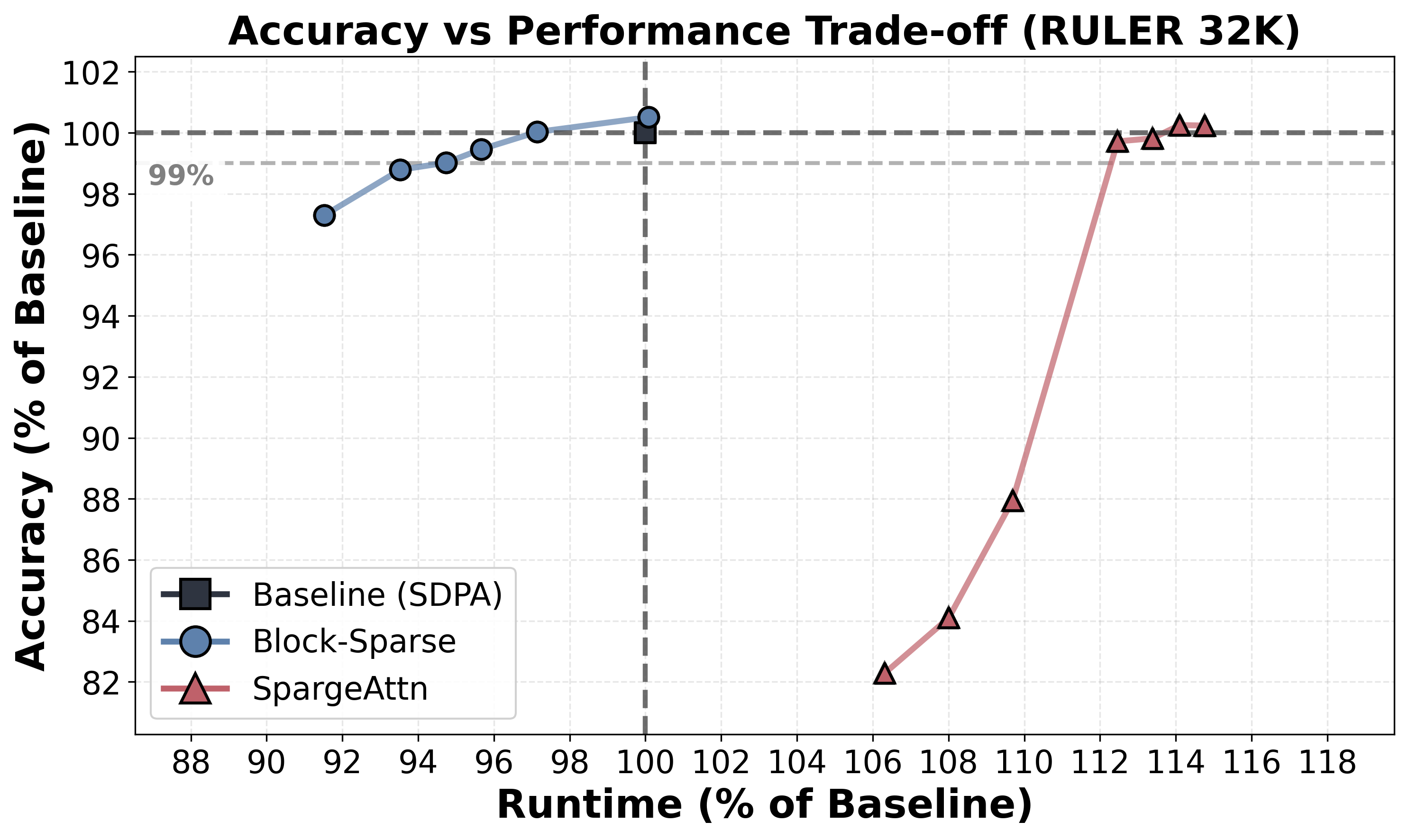}
  \caption{32K}
\end{subfigure}\hfill
\begin{subfigure}[t]{0.45\linewidth}
  \centering
  \includegraphics[width=\linewidth]{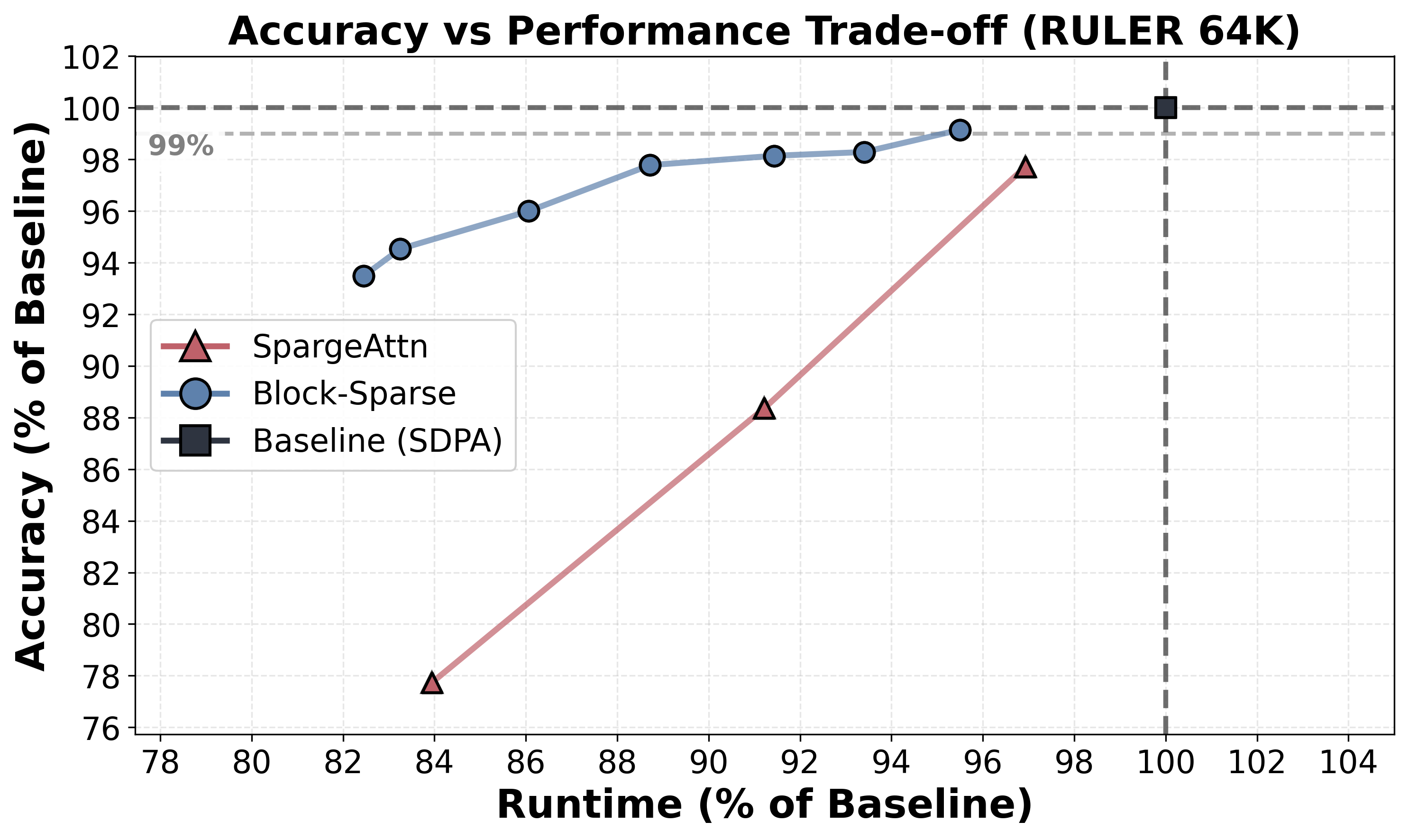}
  \caption{64K}
\end{subfigure}\\[-4pt]
\begin{subfigure}[t]{0.45\linewidth}
  \centering
  \includegraphics[width=\linewidth]{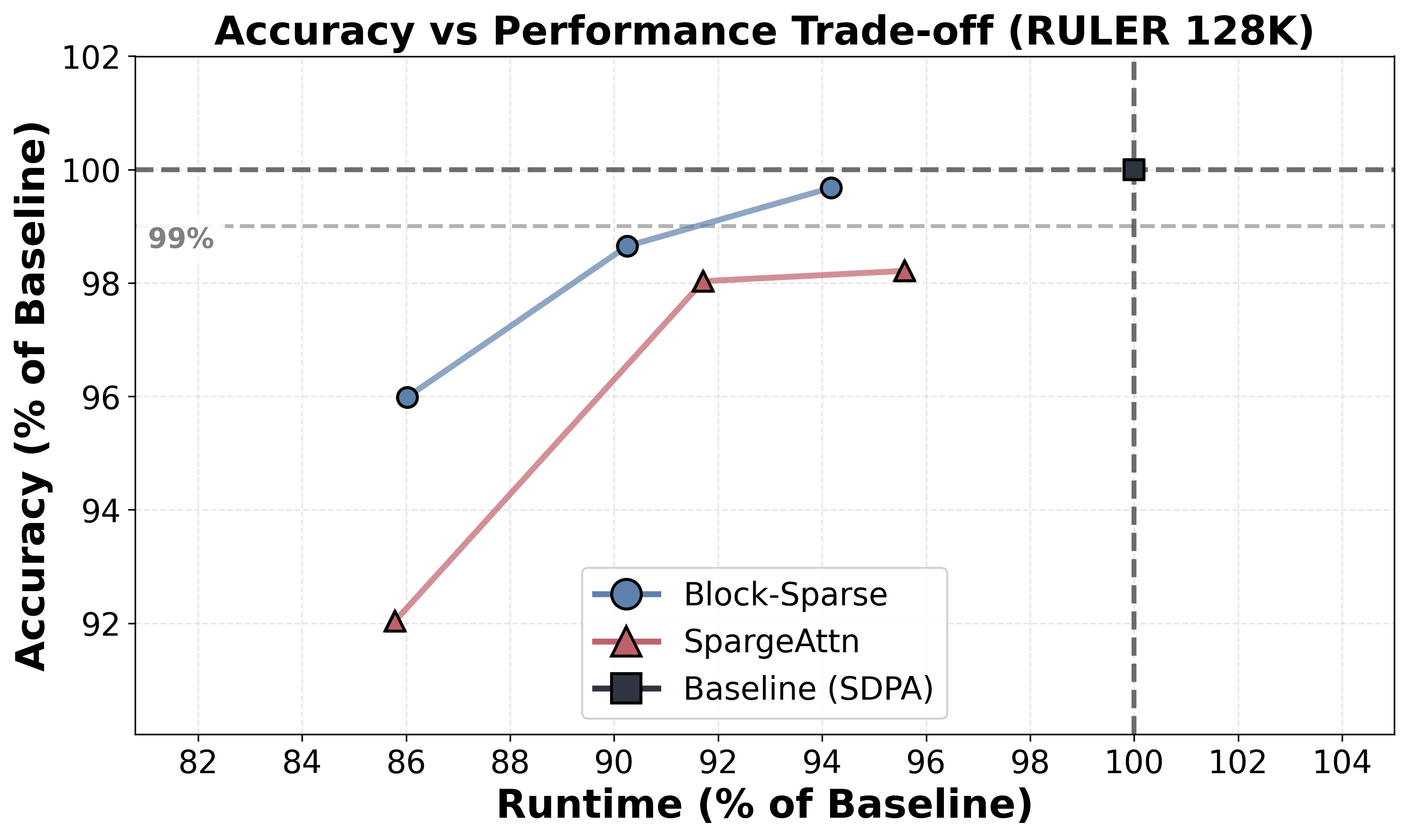}
  \caption{128K}
\end{subfigure}
\caption{Accuracy-latency trade-offs on RULER for 32K (top-left), 64K (top-right), and 128K (bottom). BSFA (blue) maintains high accuracy with consistent speedups, outperforming SpargeAttention (orange). Each point is a different sparsity level (Table~\ref{tab:full-ruler}).}
\label{fig:accuracy-latency}
\end{figure}


\section{Background}
\label{sec:background}

\paragraph{Scaled dot-product attention.} Transformers use multi-head scaled dot-product attention to process sequences \citep{vaswani2023attentionneed}, where each head computes $\text{softmax}(\mathbf{Q}\mathbf{K}^\top / \sqrt{d})\mathbf{V}$ and $\mathbf{Q}, \mathbf{K}, \mathbf{V}$ are linear projections of the input. Both the score computation $\mathbf{Q}\mathbf{K}^\top$ and value aggregation $\mathbf{P}\mathbf{V}$ scale as $O(N^2 d)$, dominating computation for long sequences. For Llama-3.1-8B at 128K tokens each of these quadratic operations alone requires ${\sim}32\times$ the FLOPs of a single linear projection (see Appendix~\ref{app:attention}).

\paragraph{FlashAttention} FlashAttention-2 (FA-2)~\citep{dao2023flashattention2fasterattentionbetter} computes exact attention without materializing the full $N \times N$ score matrix by tiling queries into $M_Q$ blocks of size $B_M$ and keys/values into $M_{KV}$ blocks of size $B_N$, keeping intermediate results in fast on-chip SRAM while streaming from HBM. For each tile $(i,j)$, FA-2 performs two equally expensive matrix multiplications (computing scores $\mathbf{S}_{ij} = \mathbf{Q}_i\mathbf{K}_j^\top$ and the value aggregation $\tilde{\mathbf{P}}_{ij}\mathbf{V}_j$, each $2B_M B_N d$ FLOPs) while loading both $\mathbf{K}_j$ and $\mathbf{V}_j$ blocks from HBM. FA-2 removes the memory bottleneck but still computes all $O(N^2)$ interactions, motivating block-sparse approaches that selectively skip negligible blocks (Appendix~\ref{app:flashattention} provides full algorithmic details).

\paragraph{Empirical sparsity patterns.}
\label{sec:attention-sparsity-patterns}
Attention distributions in transformers exhibit inherent sparsity. The phenomenon of \emph{rank collapse}~\citep{dong2023attentionneedpureattention, saada2025mind} forces attention matrices to degenerate into low-rank forms in deeper layers, concentrating mass on a small number of positions. As a result, models consistently develop \emph{attention sinks} at initial positions~\citep{xiao2024efficientstreaminglanguagemodels} and concentrate on sparse ``heavy hitter'' tokens~\citep{zhang2023h2oheavyhitteroracleefficient, zucchet2025emergencesparseattentionimpact}, while most query-key interactions contribute negligibly to the output~\citep{tang2024questqueryawaresparsityefficient}.


\section{Related Work}
\label{sec:related-work}

As transformers scale to longer contexts, various approaches have emerged to reduce attention's quadratic complexity. Since attention consists of three main operations (computing QK scores, applying softmax, and computing PV products), different methods target different subsets of these operations. We categorize existing work by which operations they optimize and when sparsity decisions are made.

\paragraph{Predicting importance before scoring.}
Methods that avoid computing $\mathbf Q\mathbf K^\top$ for certain pairs can in principle achieve the greatest speedup by eliminating all three operations for pruned blocks, but must predict importance without observing the actual attention scores. \emph{Pattern-based} approaches fix sparsity a priori using local windows with global tokens \citep{beltagy2020longformerlongdocumenttransformer}, strided patterns \citep{child2019generatinglongsequencessparse}, or block-diagonal structures \citep{gupta2021memoryefficienttransformerstopkattention}, and cannot adapt to content-specific importance. \emph{Content-aware} approaches predict importance from approximations of the score matrix. Hashing groups similar tokens \citep{kitaev2020reformerefficienttransformer}, clustering does the same with online k-means \citep{roy2020efficientcontentbasedsparseattention}, and Quest~\citep{tang2024questqueryawaresparsityefficient} uses lightweight query-aware heuristics to prune key-value pairs.

A more recent line of work targets prefill specifically. SpargeAttention~\citep{zhang2025spargeattentionaccuratetrainingfreesparse} compresses Q/K blocks via mean-pooling for a coarse global filter, and inside the kernel additionally skips $\mathbf{P}\mathbf{V}$ when the block local maximum is far below the running global maximum. MInference~\citep{jiang2024minference} assigns each attention head offline to one of three sparse patterns (A-shape, Vertical-Slash, Block-Sparse) via kernel-aware search. FlexPrefill~\citep{lai2025flexprefill} chooses per-head patterns and indices online from pooled-block and last-block approximations, controlled by a coverage threshold $\gamma$. XAttention~\citep{xu2025xattention} sums strided antidiagonals of the score block as a cheap importance proxy, selecting blocks by a cumulative-mass threshold. What unifies these methods is that pruning commits before exact scores are seen, with sparse patterns varying across queries, heads, and inputs. Variable per-query workload bottlenecks GPU kernels at the slowest thread, and the pre-attention work introduces a fixed overhead that becomes a regression on shorter sequences, a phenomenon we observe in Section~\ref{sec:experiments}.

\paragraph{Gating after scoring.}
H2O \citep{zhang2023h2oheavyhitteroracleefficient} tracks cumulative attention weights during decoding and evicts low-scoring KV cache entries. This targets decode-time memory rather than prefill compute, and evicted tokens cannot be recovered later. StreamingLLM \citep{xiao2024efficientstreaminglanguagemodels} retains attention-sink tokens plus a sliding window for streaming, but uses fixed patterns rather than content-based selection. KV-cache compression methods such as KVzip~\citep{kim2025kvzipqueryagnostickvcache} and KVzap~\citep{jegou2026kvzapfastadaptivefaithful} reduce the memory footprint of the KV cache. Neither reports a prefill or time-to-first-token speedup, the latency BSFA targets, so they complement BSFA rather than replace it.

BLASST~\citep{yuan2025blasstdynamicblockedattention}, like BSFA, gates blocks based on attention-score maxima rather than approximations, and is the closest method to ours methodologically. The two approaches differ in threshold parameterization and in comparison reference. BLASST applies a single global threshold per (model, target-sparsity) scaling with context length, while BSFA stores a per-(layer, head, position) threshold table calibrated offline that directly targets the per-layer/head heterogeneity shown in BLASST's Figure~7. BLASST also gates based on the gap to a \emph{running} global maximum (variable sparsity across queries and positions), where BSFA's fixed-$k$ design yields a uniform per-query workload that avoids the slowest-thread bottleneck. We compare against BLASST's public release on H100, since no A100 implementation is provided.

$\Delta$-Attention~\citep{willette2025delta} addresses the complementary problem of correcting distributional shift from sparse prefill via dense-sparse output differences for sampled query rows. This correction is orthogonal to block selection and could potentially be combined with BSFA.

\paragraph{Learned sparsity.}
SparseK~\citep{lou2024sparserfastermoreefficient} and SeerAttention~\citep{gao2024seerattention} learn which keys to attend to through trainable scoring networks with differentiable selection. While these yield adaptive, content-aware sparsity, they require modifying the model architecture and fine-tuning. In contrast, our method remains training-free: we calibrate thresholds offline on a small dataset without updating any model weights or introducing learnable components.

\paragraph{Approximation methods.}
Orthogonal to selection-based sparsity, kernel-based \citep{choromanski2022rethinkingattentionperformers} and low-rank \citep{xiong2021nystromformernystrombasedalgorithmapproximating} methods replace softmax attention with computationally cheaper approximations. While these offer strong asymptotic complexity guarantees, they fundamentally alter the attention mechanism and require architectural changes.


\section{Method}

\begin{figure}[t]
\centering
\includegraphics[width=\linewidth]{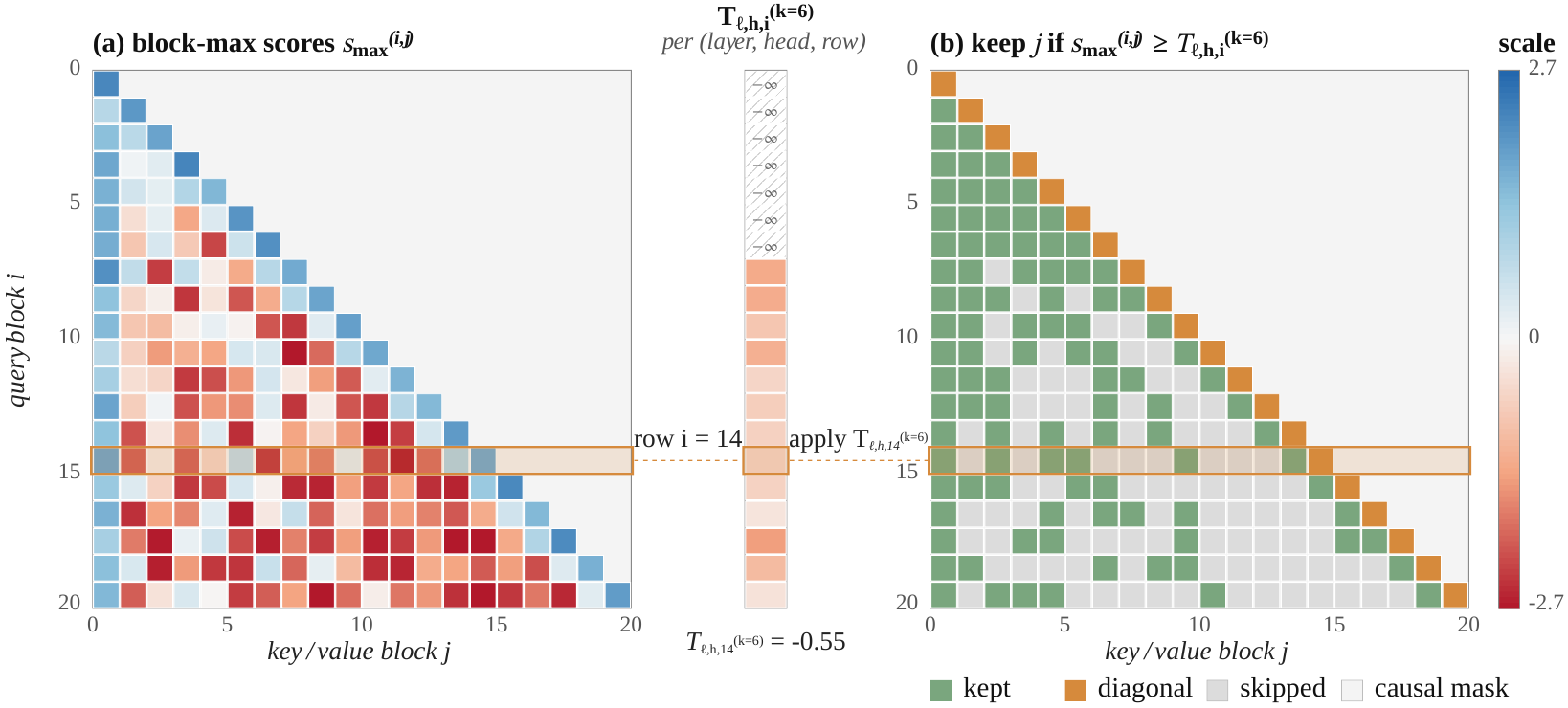}
\caption{BSFA mechanism. \textbf{(a)} Per-block maxima $s_{\max}^{(i,j)}$ are computed exactly within the FlashAttention tile. The center column visualizes the calibrated threshold tensor $T_{\ell,h,i}^{(k)}$ ($k\!=\!6$ shown), one value per query row, with hatched $-\infty$ entries where fewer than $k$ off-diagonal candidates exist. \textbf{(b)} Off-diagonal blocks are kept (green) if $s_{\max}^{(i,j)} \geq T_{\ell,h,i}^{(k)}$. Diagonal blocks (orange) always run; skipped blocks (grey) avoid the $\mathbf{V}_j$ load and the $\mathbf{PV}$ matmul. The highlighted row $i\!=\!14$ traces a single gating decision across all three panels.}
\label{fig:method-overview}
\end{figure}

We present Block Sparse Flash Attention (BSFA), which accelerates long-context prefill by selecting the top-$k$ most important value blocks for each query using calibrated thresholds (Figure~\ref{fig:method-overview}). Building on FlashAttention's~\citep{dao2023flashattention2fasterattentionbetter} tiled computation, BSFA adds a single check after each score block: if the per-block maximum score falls below a calibrated threshold specific to that layer, head, and position, the block's values are not loaded from HBM and the corresponding $\mathbf{PV}$ multiplication is skipped.

Block-level sparsity can be reliably detected through exact-score maxima rather than predicted before observing scores. Computing all $\mathbf{Q}\mathbf{K}^\top$ products preserves attention fidelity, while the skipped $\mathbf{PV}$ multiplication and value loads constitute approximately $50\%$ of attention FLOPs and $50\%$ of HBM bandwidth (Section~\ref{sec:computational-savings}). Operating at block granularity aligns with tensor cores and amortizes each gating decision over thousands of value loads. Following FlashAttention's tiling, $\mathbf{Q}, \mathbf{K}, \mathbf{V} \in \mathbb{R}^{N \times d}$ are partitioned into blocks of size $B_M$ and $B_N$, and the number of retained blocks $k$ serves as our sparsity parameter. An additional advantage of fixed-$k$ is improved GPU utilization, since variable sparsity creates workload imbalance bottlenecked by the slowest thread.

\subsection{Gating Mechanism}

Our gating mechanism determines which value blocks to process by examining the maximum attention score within each query-key block pair. For each block pair $(i,j)$ with $j<i$ in the causal setting, we compute the standard scaled dot-product scores (as defined in Section~\ref{sec:background}):
\begin{equation}
\mathbf{S}_{ij} = \frac{\mathbf{Q}_i\mathbf{K}_j^\top}{\sqrt{d}} \in \mathbb{R}^{B_M \times B_N}
\end{equation}
and extract the maximum score:
\begin{equation}
s_{\max}^{(i,j)} = \max_{p,q} [\mathbf{S}_{ij}]_{pq}
\end{equation}

We select the top-$k$ blocks per query using thresholds calibrated per (layer, head, position). Unlike SpargeAttention, which skips blocks whose post-softmax contributions become numerically negligible after exponential scaling, BSFA targets a fixed number of retained blocks for consistent workload across inputs. We gate each block by comparing $s_{\max}^{(i,j)}$ against $T_{\ell,h,i}^{(k)}$ (Figure~\ref{fig:method-overview}b). If below, the block is skipped entirely: values are not loaded, the $\mathbf{PV}$ multiplication is not performed, and running statistics are not updated. The threshold parameterization over $(\ell, h, i, k)$ captures attention's empirical heterogeneity across layers (local in lower, sink-attentive in deeper), heads (local vs.\ global specialists), and positions~\citep{xiao2024efficientstreaminglanguagemodels}. Diagonal blocks bypass the gate (Figure~\ref{fig:method-overview}b, orange) to preserve causal correctness and local context.

\subsection{Threshold Calibration}
\label{sec:threshold-calibration}

A key observation underlying our approach is that attention patterns remain relatively consistent across different inputs for a given model, allowing offline calibration on a small dataset. The procedure is simple: for each query position, we sort off-diagonal blocks by their maximum scores and select the threshold that retains exactly the top-$k$ blocks, averaging across calibration samples for robustness. We store thresholds per layer, head, and position ($\mathbf{T} \in \mathbb{R}^{L \times H \times \lceil N_{\max}/B_M \rceil}$), with multiple sparsity levels enabling dynamic accuracy-speed trade-offs at deployment without recalibration (see Appendix~\ref{app:calibration} for details).

\subsection{Complexity Analysis}
\label{sec:computational-savings}

By skipping blocks with low maximum scores, BSFA eliminates substantial computation and memory traffic. For each skipped block, we avoid:

\textbf{Compute operations:} The dominant saving comes from skipping the $\mathbf{PV}$ multiplication, which requires $2B_M B_N d$ FLOPs per block, the same cost as the $\mathbf{QK}$ computation. We also skip exponential computations for softmax ($B_M B_N$ operations), running statistics updates ($O(B_M d)$ operations), and normalizer maintenance ($O(B_M B_N)$ operations).

\textbf{Memory bandwidth:} We avoid loading value blocks from HBM, saving $B_N d$ elements per skipped block. Since keys and values are the same size, this represents exactly 50\% of the key-value memory traffic.

The overhead of computing block maxima and threshold comparisons is negligible compared to the tensor operations we skip.

\subsection{Algorithm and Implementation}
Algorithm~\ref{alg:gated-fa-prefill} presents our block sparse Flash Attention for the causal prefill case. The modification to standard FlashAttention is minimal: we add a gating check after computing each score block but before loading values. The integration requires only a conditional branch in the existing kernel: if $s_{\max}^{(i,j)} < T_{\ell,h,i}^{(k)}$ (where the threshold is specific to layer $\ell$, head $h$, and position $i$), we skip value loading and all subsequent operations for that block. Skipped blocks do not participate in the running statistics $(m_i, \ell_i, \mathbf{O}_i)$ maintained by the streaming softmax algorithm.

\begin{algorithm}[h!]
\caption{Block Sparse Flash Attention (simplified for single query block $i$, causal)}
\label{alg:gated-fa-prefill}
\begin{algorithmic}[1]
\STATE \textbf{Input:} Query block $\mathbf{Q}_i \in \mathbb{R}^{B_M \times d}$, Key blocks $\{\mathbf{K}_j\}$, Value blocks $\{\mathbf{V}_j\}$, Thresholds $T_{\ell,h,i}^{(k)}$
\STATE \textbf{Initialize:} $m_i = -\infty \cdot \mathbf{1}_{B_M}$, $\ell_i = \mathbf{0}_{B_M}$, $\tilde{\mathbf{O}}_i = \mathbf{0}_{B_M \times d}$ \COMMENT{un-scaled accumulator}
\STATE Load $\mathbf{Q}_i$ from HBM to SRAM
\FOR{each key/value block $j = 0, \ldots, i-1$}
    \STATE Load $\mathbf{K}_j$ from HBM to SRAM
    \STATE Compute scores: $\mathbf{S}_{ij} = \mathbf{Q}_i\mathbf{K}_j^\top / \sqrt{d}$ \COMMENT{in SRAM}
    \STATE Compute block maximum: $s_{\max}^{(i,j)} = \max(\mathbf{S}_{ij})$
    \IF{$s_{\max}^{(i,j)} > T_{\ell,h,i}^{(k)}$}
        \STATE Update row-wise max: $m_i^{\text{new}} = \max(m_i, \text{rowmax}(\mathbf{S}_{ij}))$
        \STATE Compute $\tilde{\mathbf{P}}_{ij} = \exp(\mathbf{S}_{ij} - m_i^{\text{new}})$ \COMMENT{stable softmax}
        \STATE Load $\mathbf{V}_j$ from HBM to SRAM
        \STATE Rescale accumulator: $\tilde{\mathbf{O}}_i \leftarrow \tilde{\mathbf{O}}_i \cdot \text{diag}(\exp(m_i - m_i^{\text{new}}))$
        \STATE Update un-scaled output: $\tilde{\mathbf{O}}_i \leftarrow \tilde{\mathbf{O}}_i + \tilde{\mathbf{P}}_{ij}\mathbf{V}_j$
        \STATE Update normalizer: $\ell_i \leftarrow \ell_i \cdot \exp(m_i - m_i^{\text{new}}) + \text{rowsum}(\tilde{\mathbf{P}}_{ij})$
        \STATE $m_i \leftarrow m_i^{\text{new}}$
    \ENDIF
\ENDFOR
\STATE Process diagonal block $j = i$ with causal mask applied to $\mathbf{S}_{ii}$ \COMMENT{always process, no gating}
\STATE \textbf{Finalize:} $\mathbf{O}_i \leftarrow \tilde{\mathbf{O}}_i \oslash \ell_i$ \COMMENT{final scaling, element-wise division}
\STATE Store $L_i = m_i + \log(\ell_i)$ \COMMENT{logsumexp for backward pass}
\STATE \textbf{Return:} $\mathbf{O}_i$
\end{algorithmic}
\end{algorithm}


\section{Experimental Evaluation}
\label{sec:experiments}

We evaluate Block Sparse Flash Attention (BSFA) on diverse long-context benchmarks to demonstrate its effectiveness in reducing prefill latency while maintaining model quality. Our experiments span multiple sequence lengths, two language-model architectures, three GPU architectures, and five recent sparse-attention baselines.

\subsection{Experimental Setup}

\paragraph{Hardware and implementation.}
Our primary experiments are conducted on NVIDIA A100 80GB GPUs with CUDA 12.1. We implemented BSFA as a custom CUDA kernel extending FlashAttention~2~\citep{dao2023flashattention2fasterattentionbetter}, with block dimensions $B_M = 128$ and $B_N = 64$ optimized for A100's tensor cores. The kernel modifications add threshold-based gating logic with minimal overhead. BSFA uses FP16 precision throughout, maintaining numerical fidelity while achieving speedups through selective computation. To verify generality, we additionally evaluate on RTX A6000 (Ampere, lower memory bandwidth) and on H100 with a Hopper-compatible kernel port. Details are deferred to the generality discussion in \S\ref{sec:experiments} and Appendix~\ref{app:hardware}. Our complete implementation is available at \url{https://github.com/Danielohayon/Block-Sparse-Flash-Attention}, and all reported results are fully reproducible using the provided scripts.

\paragraph{Models and datasets.}
We primarily evaluate on Llama-3.1-8B-Instruct~\citep{grattafiori2024llama3herdmodels}, supplementing with cross-model evaluation on Qwen2.5-7B-Instruct~\citep{qwen2025qwen25technicalreport} (28 layers, 28 heads, 4 KV heads, GQA-7) to verify that the method generalizes to a different architecture. We use three benchmarks. \textbf{LongBench}~\citep{bai2024longbenchbilingualmultitaskbenchmark} is a real-world long-document understanding benchmark spanning multiple domains. We report accuracy on the full 27-task benchmark ($\sim$4{,}750 samples) and TTFT on two subsets, the 10 longest \texttt{narrativeqa} samples ($\sim$65K tokens each) and a mixed-length subset of 160 samples spanning 295--65{,}461 tokens for the variable-length analysis. \textbf{RULER}~\citep{hsieh2024rulerwhatsrealcontext} provides synthetic tasks testing precise retrieval and reasoning over long contexts (32K, 64K, 128K tokens). \textbf{Needle-in-a-Haystack} provides targeted retrieval at 64K.

Crucially, the calibration and evaluation datasets are disjoint across task categories. RULER contains multiple task categories (Retrieval, Multi-hop Tracing, Aggregation, Question Answering, etc.), which we split into two groups. For threshold calibration, we use 16 samples from the first group of categories. For RULER latency measurement, we select samples per sequence length exclusively from the held-out categories not used during calibration, ensuring our speedup measurements reflect generalization to unseen task types. TTFT exhibits low variance across samples of similar length, making 10 samples sufficient for reliable timing in our primary measurements. Per-experiment sample counts are noted in their respective tables. We use the same RULER-calibrated thresholds for our LongBench evaluation (Table~\ref{tab:longbench-main}), demonstrating strong cross-dataset generalization without any dataset-specific tuning. For accuracy evaluation, we test the full RULER and LongBench benchmarks using the lm-evaluation-harness framework~\citep{eval-harness}.

\paragraph{Baselines.}
Our exact-attention reference is \textbf{Dense FlashAttention-2}~\citep{dao2023flashattention2fasterattentionbetter} (we use FA-2 since FA-3 is optimized for Hopper and not available on A100). On the sparse-attention side we evaluate \textbf{SpargeAttention}~\citep{zhang2025spargeattentionaccuratetrainingfreesparse}, \textbf{MInference}~\citep{jiang2024minference}, \textbf{FlexPrefill}~\citep{lai2025flexprefill}, \textbf{XAttention}~\citep{xu2025xattention}, and \textbf{BLASST}~\citep{yuan2025blasstdynamicblockedattention}, all described in Section~\ref{sec:related-work}. We follow each method's recommended configuration and sweep their natural accuracy-speed knobs (SpargeAttention's $\tau$, FlexPrefill's $\gamma$, BLASST's sparsity target). Since BLASST has no A100 release, we run it against their public H100 release. We additionally include a \textbf{Sliding Window} baseline at the same block budget as BSFA, to confirm content-aware selection does not simply collapse to a local pattern. Method-specific implementation choices (kernel selection, dtype constraints, and other artifact-specific notes) are documented in Appendix~\ref{app:baselines}.

\phantomsection
\paragraph{Metrics.}
\label{par:metrics}
We measure three quantities. \textbf{End-to-end TTFT} is the wall-clock prefill time, averaged over 10 runs after warmup. \textbf{Attention-kernel-only TTFT} isolates the attention call via CUDA-event timing around the kernel boundary, allowing us to disambiguate kernel-level speedup from end-to-end speedup that includes work BSFA cannot accelerate (Q/K/V/O projections, RoPE, GQA expansion, MLP, the LM head). \textbf{Task accuracy} is evaluated using lm-evaluation-harness~\citep{eval-harness}. We additionally validate threshold effectiveness by comparing predicted versus measured block density. For query block $i$,
\begin{equation}
\rho_i = \frac{\min(k, i) + 1}{i + 1},
\label{eq:density}
\end{equation}
where $k$ is the number of retained off-diagonal blocks and $+1$ accounts for the always-retained diagonal block. Measured density is the actual fraction retained on validation data. We measure both accuracy and latency on the \emph{same} sequence lengths and samples for each sparsity configuration.


\subsection{Main Results}

\begin{table}[t]
\centering
\caption{Main results on LongBench (Llama-3.1-8B). Accuracy is on the full 27-task benchmark ($\sim$4{,}750 samples) via lm-evaluation-harness, and speedup is end-to-end TTFT on the 10 longest \texttt{narrativeqa} samples ($\sim$65K tokens each). All methods on A100 unless marked $\star$ (H100, since BLASST has no public A100 implementation). Sliding Window does not run as a sparse-attention kernel against the same baseline, so we report accuracy only. Bracketed numbers show relative degradation from the dense baseline.}
\label{tab:longbench-main}
\footnotesize
\setlength{\tabcolsep}{4pt}
\begin{tabular}{llcc}
\toprule
Method & Param. & Accuracy $\uparrow$ & Speedup $\uparrow$ \\
\midrule
Dense FlashAttention-2 & All & 40.24\% & 1.00$\times$ \\
\cmidrule{1-4}
\multirow{4}{*}{BSFA}
    & $k$=32 & 39.39\% (-2.1\%) & 1.16$\times$ \\
    & $k$=64 & 39.78\% (-1.1\%) & 1.13$\times$ \\
    & $k$=96 & 39.88\% (-0.9\%) & 1.10$\times$ \\
    & $k$=128 & 40.03\% (-0.5\%) & 1.08$\times$ \\
\cmidrule{1-4}
\multirow{3}{*}{SpargeAttention}
    & $\tau$=0.4 & 30.06\% (-25.3\%) & 1.09$\times$ \\
    & $\tau$=0.5 & 33.11\% (-17.7\%) & 1.02$\times$ \\
    & $\tau$=0.6 & 35.87\% (-10.9\%) & 0.98$\times$ \\
\cmidrule{1-4}
MInference & default & 40.06\% (-0.45\%) & 1.17$\times$ \\
\cmidrule{1-4}
\multirow{2}{*}{FlexPrefill}
    & $\gamma$=0.9 & 35.11\% (-12.8\%) & 1.59$\times$ \\
    & $\gamma$=0.99 & 38.90\% (-3.3\%) & 1.06$\times$ \\
\cmidrule{1-4}
XAttention & default & 39.95\% (-0.7\%) & 1.16$\times$ \\
\cmidrule{1-4}
\multirow{2}{*}{BLASST$^{\star}$}
    & 50\% & 39.23\% (-2.5\%) & 1.11$\times$ \\
    & 70\% & 36.40\% (-9.5\%) & 1.16$\times$ \\
\cmidrule{1-4}
\multirow{2}{*}{Sliding Window}
    & 64 (4{,}096 tokens) & 13.43\% (-66.6\%) & --- \\
    & 96 (6{,}144 tokens) & 20.29\% (-49.6\%) & --- \\
\bottomrule
\end{tabular}
\end{table}

\paragraph{LongBench accuracy and speedup.}
Table~\ref{tab:longbench-main} presents the head-to-head comparison. BSFA's $k$ slider exposes a flexible operating range from $1.08\times$/$-0.5\%$ ($k$=128) to $1.16\times$/$-2.1\%$ ($k$=32), all served from a single calibrated threshold tensor. MInference operates at a fixed point ($1.17\times$/$-0.45\%$) chosen offline by pattern search, with no accuracy-speed knob and a per-task spread from $-6.0\%$ on hotpotqa to $+9.3\%$ on passage\_count, plus a pre-attention overhead regression to $0.17\times$ at $<\!4$K (Table~\ref{tab:variable-length}). XAttention reaches $1.16\times$ at $-0.7\%$ with a milder overhead floor ($0.85\times$ at $<\!4$K). BLASST trades accuracy more aggressively at matched speedup, reaching $1.11\times$ at $50\%$ sparsity but losing $2.5\%$ versus BSFA $k$=64 at $1.13\times$ with only $1.1\%$ loss. SpargeAttention loses at least $10.9\%$ and FlexPrefill at its best-quality $\gamma$=0.99 still loses $3.3\%$ at $1.06\times$, neither reaching the $99\%$-accuracy band. The sliding-window baseline at the same block budget collapses to $\le\!21\%$ accuracy, confirming that content-aware selection is essential.

\begin{table}[t]
\centering
\caption{End-to-end TTFT speedup on a LongBench mixed-length subset (160 samples spanning 295--65{,}461 tokens), broken down by sequence-length bucket. Speedup is reported against dense FlashAttention-2 within each bucket. Accuracy column is the full 27-task LongBench score from Table~\ref{tab:longbench-main}, repeated here for ease of speedup-vs-accuracy correlation. Llama-3.1-8B on A100 unless marked $\star$ (H100). SpargeAttention was not measured on this subset. \textbf{Bold} cells highlight regressions below dense ($<\!1.00\times$).}
\label{tab:variable-length}
\footnotesize
\setlength{\tabcolsep}{3pt}
\begin{tabular}{llcccccc}
\toprule
Method & Param. & Accuracy $\uparrow$ & $<$4K & 4--8K & 8--16K & 16--32K & 32--65K \\
\midrule
Dense FlashAttention-2 & All & 40.24\% & 1.00$\times$ & 1.00$\times$ & 1.00$\times$ & 1.00$\times$ & 1.00$\times$ \\
\cmidrule{1-8}
BSFA & $k$=64 & 39.78\% (-1.1\%) & \textbf{0.96$\times$} & \textbf{0.97$\times$} & 1.01$\times$ & 1.05$\times$ & 1.19$\times$ \\
\cmidrule{1-8}
MInference & default & 40.06\% (-0.45\%) & \textbf{0.17$\times$} & \textbf{0.31$\times$} & \textbf{0.48$\times$} & \textbf{0.58$\times$} & \textbf{0.99$\times$} \\
\cmidrule{1-8}
\multirow{2}{*}{FlexPrefill}
    & $\gamma$=0.9  & 35.11\% (-12.8\%) & \textbf{0.71$\times$} & \textbf{0.90$\times$} & 1.05$\times$ & 1.18$\times$ & 1.59$\times$ \\
    & $\gamma$=0.99 & 38.90\% (-3.3\%) & \textbf{0.69$\times$} & \textbf{0.80$\times$} & \textbf{0.93$\times$} & \textbf{0.99$\times$} & 1.12$\times$ \\
\cmidrule{1-8}
XAttention & default & 39.95\% (-0.7\%) & \textbf{0.85$\times$} & \textbf{0.90$\times$} & \textbf{0.97$\times$} & 1.01$\times$ & 1.09$\times$ \\
\cmidrule{1-8}
\multirow{2}{*}{BLASST$^{\star}$}
    & 50\% & 39.23\% (-2.5\%) & \textbf{0.97$\times$} & 1.01$\times$ & 1.03$\times$ & 1.06$\times$ & 1.08$\times$ \\
    & 70\% & 36.40\% (-9.5\%) & 1.00$\times$ & 1.02$\times$ & 1.05$\times$ & 1.09$\times$ & 1.11$\times$ \\
\bottomrule
\end{tabular}
\end{table}

\paragraph{Variable-length analysis.}
Table~\ref{tab:variable-length} reports per-bucket TTFT on the mixed-length subset. Pre-attention selection methods incur a fixed overhead per call and regress on shorter contexts (MInference $0.17\times$ below 4K, FlexPrefill $\gamma$=0.99 $0.69\times$, XAttention $0.85\times$). BSFA's gate is one block-max comparison inline in the FA-2 kernel with essentially no overhead, mildly negative ($\sim\!0.96\times$) below 8K and positive at $\geq$8K. BLASST on H100 is Hopper-only and loses more accuracy at matched sparsity (Table~\ref{tab:longbench-main}).

\paragraph{RULER 64K and the full sweep.}
BSFA's $k$ provides a continuous accuracy-speedup knob. Figure~\ref{fig:accuracy-latency} traces this curve against SpargeAttention at RULER 32K, 64K, and 128K, with BSFA outperforming SpargeAttention's full sweep at every context length. The same calibrated thresholds let BSFA cover the high-accuracy regime on LongBench ($k$=96 at $1.10\times$/$99.1\%$, Table~\ref{tab:longbench-main}), an operating point FlexPrefill ($-3.3\%$) and BLASST ($-2.5\%$) cannot reach at any evaluated configuration. At more aggressive sparsity, BSFA's $k$=192 on RULER 64K is competitive in speedup with MInference, FlexPrefill at $\gamma$=0.99, and BLASST 50\% on H100 (full comparison in Appendix Table~\ref{tab:ruler-64k}).

\paragraph{Attention-only versus end-to-end speedup.}
End-to-end TTFT includes work BSFA cannot accelerate, such as Q/K/V/O projections, RoPE, GQA expansion, MLP, and the LM head. Table~\ref{tab:attention-only} isolates the attention kernel via CUDA-event timing around the kernel boundary. At RULER 64K $k$=192, the attention kernel itself is $1.38\times$ faster, and amortizes to $1.13\times$ over the full forward pass. The kernel-only number is the apples-to-apples FlashAttention-replacement metric, while the end-to-end number is what users observe when BSFA is dropped into a full inference pipeline.

\begin{table}[!ht]
\centering
\caption{Attention-kernel-only versus end-to-end speedup at RULER 64K (BSFA, Llama-3.1-8B, A100). Accuracy is on RULER 64K with relative degradation from dense in brackets. Kernel-only timing isolates the attention call from work BSFA cannot accelerate (Q/K/V/O projections, RoPE, GQA expansion, MLP, the LM head).}
\label{tab:attention-only}
\footnotesize
\setlength{\tabcolsep}{6pt}
\begin{tabular}{lccc}
\toprule
Param. & Accuracy $\uparrow$ & E2E Speedup $\uparrow$ & Attn-only Speedup $\uparrow$ \\
\midrule
SDPA & 84.96\% & 1.00$\times$ & 1.00$\times$ \\
$k$=192 & 83.08\% (-2.2\%) & 1.13$\times$ & \textbf{1.38$\times$} \\
$k$=256 & 83.39\% (-1.8\%) & 1.09$\times$ & 1.30$\times$ \\
$k$=384 & 84.24\% (-0.8\%) & 1.05$\times$ & 1.19$\times$ \\
\bottomrule
\end{tabular}
\end{table}

\paragraph{Task-adaptive sparsity (Needle-in-a-Haystack).}

\begin{figure}[!htb]
\centering
\includegraphics[width=0.6\linewidth,keepaspectratio]{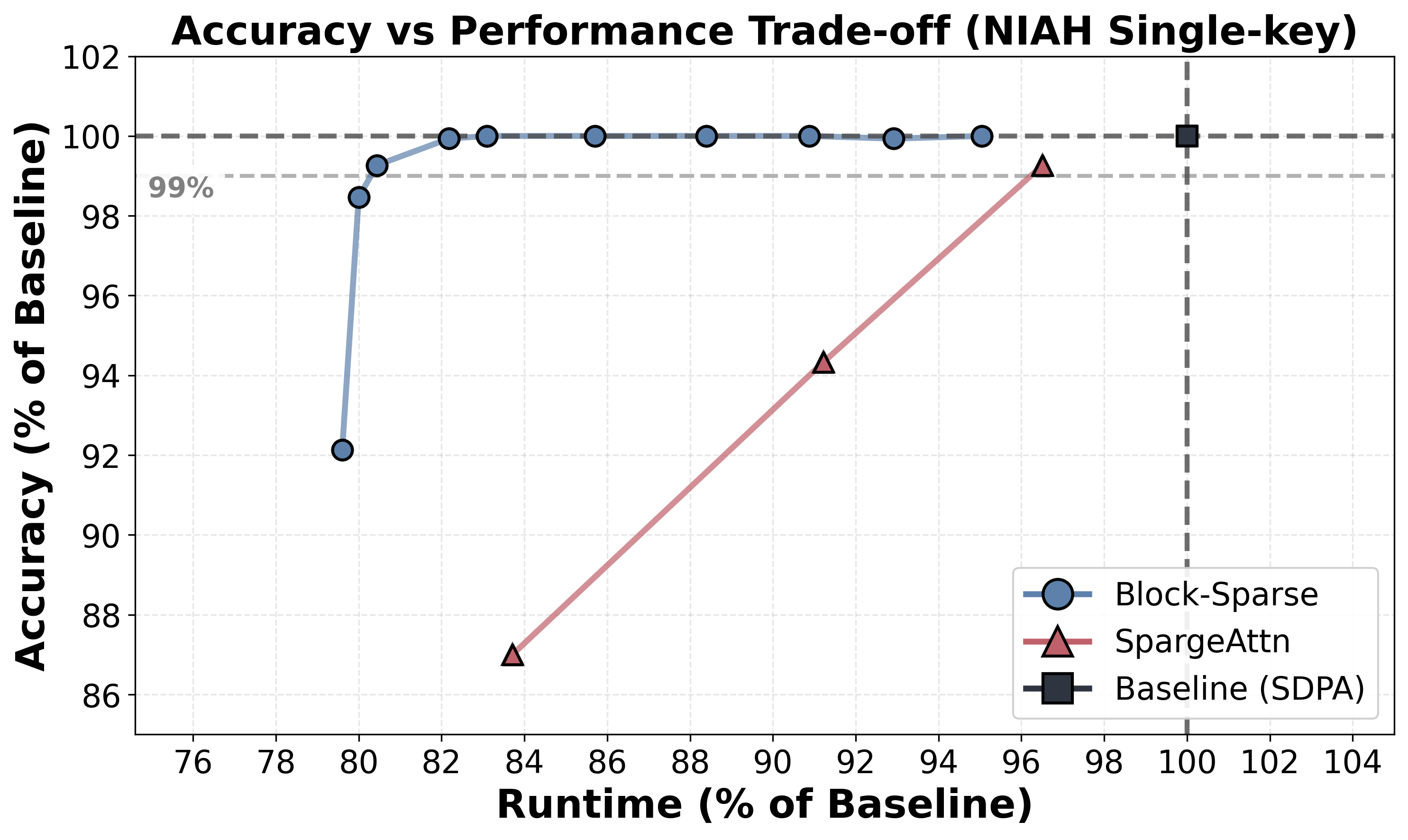}
\caption{Accuracy vs.\ Time-to-First-Token (TTFT) speedup on the 64K Needle-in-a-Haystack Single-key task. BSFA retains 99\% of baseline accuracy at $k{=}32$ with a $1.24\times$ speedup, exploiting that retrieval concentrates signal in few blocks. SpargeAttention is shown as a multi-operating-point reference. The remaining baselines (MInference, FlexPrefill, XAttention, BLASST) were not evaluated on this task and are compared on LongBench in Tables~\ref{tab:longbench-main}--\ref{tab:variable-length}.}
\label{fig:niah-results}
\end{figure}

BSFA's content-aware sparsity particularly excels at targeted retrieval. Figure~\ref{fig:niah-results} presents the 64K Needle-in-a-Haystack Single-key task, where BSFA maintains 99\% of baseline accuracy with only $k$=32 blocks at $1.24\times$ speedup, a much smaller block budget than the high-accuracy LongBench operating points ($k\!\geq\!96$) in Table~\ref{tab:longbench-main}. The disparity is task-adaptive. Needle-finding concentrates critical information in a few specific locations that exact-score gating reliably identifies, while haystack distractors receive uniformly low scores. Tasks requiring broader context (summarization, multi-hop reasoning) need more blocks to capture distributed information.

\begin{table}[!ht]
\centering
\begin{minipage}{0.49\linewidth}
\centering
\caption{BSFA on additional GPU architectures (Llama-3.1-8B). A6000 reuses A100-calibrated thresholds \emph{without recalibration} on RULER 64K. H100 uses a Hopper-compatible kernel against an FA-2-style baseline on LongBench longest-10.}
\label{tab:extra-hardware}
\footnotesize
\setlength{\tabcolsep}{4pt}
\begin{tabular}{lccc}
\toprule
GPU & $k$=32 & $k$=64 & $k$=128 \\
\midrule
A6000 & 1.31$\times$ & 1.28$\times$ & 1.23$\times$ \\
H100  & 1.25$\times$ & 1.21$\times$ & 1.12$\times$ \\
\bottomrule
\end{tabular}
\vspace{6pt}
\end{minipage}\hfill
\begin{minipage}{0.49\linewidth}
\centering
\caption{BSFA on Qwen2.5-7B (A100, GQA-7, 28L/28H/4 KV-heads, separately calibrated thresholds). Accuracy is on full LongBench, and speedup is end-to-end TTFT on RULER 64K.}
\label{tab:qwen}
\footnotesize
\setlength{\tabcolsep}{4pt}
\begin{tabular}{lcc}
\toprule
Param. & Accuracy $\uparrow$ & Speedup $\uparrow$ \\
\midrule
Dense FA-2 & 34.05\% & 1.00$\times$ \\
$k$=16 & 33.17\% (-2.6\%) & 1.17$\times$ \\
$k$=32 & 33.80\% (-0.7\%) & 1.18$\times$ \\
$k$=64 & 34.06\% (+0.0\%) & 1.18$\times$ \\
\bottomrule
\end{tabular}
\end{minipage}
\end{table}

\paragraph{Generalization across hardware, models, and datasets.}
\label{sec:generality}

Tables~\ref{tab:extra-hardware} and \ref{tab:qwen} confirm BSFA transfers to A6000 (with A100-calibrated thresholds reused \emph{without recalibration}), to H100 (FA-2-style port), and to Qwen2.5-7B (matching its dense LongBench baseline at $k$=64). RULER-calibrated thresholds also drive our LongBench results (Table~\ref{tab:longbench-main}, including the $\sim$12\% Chinese portion), with measured density tracking Eq.~\ref{eq:density} (std $\le 0.06$, Appendix Table~\ref{tab:full-ruler}). Across these benchmarks, BSFA uniquely satisfies both targets, $\geq$99\% LongBench accuracy and $\geq\!0.96\times$ end-to-end speedup at every length bucket. MInference, FlexPrefill, and XAttention regress on contexts below 16K, and BLASST loses more accuracy at matched sparsity.

\paragraph{Limitations and conclusion.}
BSFA targets prefill (decode is complementary~\citep{zhang2023h2oheavyhitteroracleefficient,xiao2024efficientstreaminglanguagemodels}) and extends FA-2, leaving FA-3 integration to future work. BSFA is a training-free FA-2 replacement using calibrated per-(layer,head,position) thresholds on exact scores, validated across datasets, languages, models (Llama-3.1, Qwen2.5), and GPUs (A100, A6000, H100).

\begin{ack}
We thank the anonymous reviewers for their feedback and Habana Labs (Intel) for their support during the early stages of this work. This research was supported by the Israel Science Foundation (grant no.~1449/23) and the Pazy Research Foundation.
\end{ack}


\bibliography{references}
\bibliographystyle{plainnat}


\appendix

\section{Additional Experimental Results}
\label{app:experiments}

\subsection{Full RULER Results}
\label{app:full-ruler}

Table~\ref{tab:full-ruler} reports accuracy, predicted/measured block density, and end-to-end TTFT speedup for BSFA, SpargeAttention, and Dense FlashAttention-2 across all three RULER lengths (32K, 64K, 128K). The 64K speedup column is the source for the BSFA and Sparge $\tau$=0.5 rows in Table~\ref{tab:ruler-64k}. The Pred./Meas.\ Density columns validate the threshold calibration: measured densities track predictions (Eq.~\ref{eq:density}) with standard deviation $\le 0.06$ across all configurations and lengths.

\begin{table}[h!]
\centering
\caption{Full RULER results across multiple sparsity levels. Accuracy column shows absolute accuracy with relative degradation from baseline in brackets. Pred./Meas. Density shows predicted (Eq.~\ref{eq:density}) vs. measured density during inference (predicted/measured$\pm$std). Speedup is end-to-end TTFT vs. dense FlashAttention-2 on the same samples.}
\label{tab:full-ruler}
\footnotesize
\setlength{\tabcolsep}{3pt}
\begin{tabular}{llcccc}
\toprule
Seq Length & Method & Sparsity Param. & Accuracy $\uparrow$ & Pred./Meas. Density & Speedup $\uparrow$ \\
\midrule
\multirow{8}{*}{32K}
    & Dense FlashAttention-2 & All & 85.98\% & 1.00 & 1.00$\times$ \\
\cmidrule{2-6}
    & \multirow{4}{*}{BSFA} & $k$=64 & 84.94\% (-1.2\%) & 0.24/0.28$\pm$0.04 & 1.07$\times$ \\
    &                        & $k$=96 & 85.52\% (-0.5\%) & 0.35/0.38$\pm$0.05 & 1.04$\times$ \\
    &                        & $k$=128 & 86.00\% (+0.0\%) & 0.45/0.47$\pm$0.05 & 1.03$\times$ \\
    &                        & $k$=192 & 86.42\% (+0.5\%) & 0.62/0.62$\pm$0.05 & 1.00$\times$ \\
\cmidrule{2-6}
    & \multirow{3}{*}{SpargeAttention} & $\tau$=0.6 & 75.62\% (-12.1\%) & - & 0.91$\times$ \\
    &                                   & $\tau$=0.75 & 85.83\% (-0.2\%) & - & 0.88$\times$ \\
    &                                   & $\tau$=0.85 & 86.18\% (+0.2\%) & - & 0.87$\times$ \\
\midrule
\multirow{7}{*}{64K}
    & Dense FlashAttention-2 & All & 84.96\% & 1.00 & 1.00$\times$ \\
\cmidrule{2-6}
    & \multirow{3}{*}{BSFA} & $k$=192 & 83.08\% (-2.2\%) & 0.35/0.36$\pm$0.05 & 1.13$\times$ \\
    &                        & $k$=256 & 83.39\% (-1.8\%) & 0.44/0.45$\pm$0.05 & 1.09$\times$ \\
    &                        & $k$=384 & 84.24\% (-0.8\%) & 0.62/0.61$\pm$0.05 & 1.05$\times$ \\
\cmidrule{2-6}
    & \multirow{3}{*}{SpargeAttention} & $\tau$=0.3 & 66.04\% (-22.3\%) & - & 1.19$\times$ \\
    &                                   & $\tau$=0.4 & 75.07\% (-11.6\%) & - & 1.10$\times$ \\
    &                                   & $\tau$=0.5 & 83.03\% (-2.3\%) & - & 1.03$\times$ \\
\midrule
\multirow{6}{*}{128K}
    & Dense FlashAttention-2 & All & 76.79\% & 1.00 & 1.00$\times$ \\
\cmidrule{2-6}
    & \multirow{3}{*}{BSFA} & $k$=512 & 73.71\% (-4.0\%) & 0.44/0.44$\pm$0.06 & 1.16$\times$ \\
    &                        & $k$=768 & 75.75\% (-1.4\%) & 0.61/0.61$\pm$0.05 & 1.11$\times$ \\
    &                        & $k$=1024 & 76.54\% (-0.3\%) & 0.76/0.74$\pm$0.05 & 1.06$\times$ \\
\cmidrule{2-6}
    & \multirow{3}{*}{SpargeAttention} & $\tau$=0.5 & 70.68\% (-8.0\%) & - & 1.17$\times$ \\
    &                                   & $\tau$=0.6 & 75.28\% (-2.0\%) & - & 1.09$\times$ \\
    &                                   & $\tau$=0.7 & 75.42\% (-1.8\%) & - & 1.05$\times$ \\
\bottomrule
\end{tabular}
\end{table}

\subsection{RULER 64K Speedup, All Methods}
\label{app:ruler-64k-speedup}

Table~\ref{tab:ruler-64k} reports end-to-end TTFT speedup at the canonical RULER 64K cross-section across all methods. RULER accuracy and density are reported in Appendix Table~\ref{tab:full-ruler} for BSFA and SpargeAttention. The new sparse-attention baselines (MInference, FlexPrefill, BLASST) were not evaluated for RULER accuracy in our experiments.

\begin{table}[h!]
\centering
\caption{End-to-end TTFT speedup on RULER 64K (Llama-3.1-8B). All methods on A100 unless marked $\star$ (H100). XAttention was not measured on RULER.}
\label{tab:ruler-64k}
\footnotesize
\setlength{\tabcolsep}{4pt}
\begin{tabular}{llc}
\toprule
Method & Param. & Speedup $\uparrow$ \\
\midrule
Dense FA-2 & All & 1.00$\times$ \\
\cmidrule{1-3}
\multirow{3}{*}{BSFA}
    & $k$=192 & 1.13$\times$ \\
    & $k$=256 & 1.09$\times$ \\
    & $k$=384 & 1.05$\times$ \\
\cmidrule{1-3}
SpargeAttention & $\tau$=0.5 & 1.03$\times$ \\
MInference & default & 1.15$\times$ \\
\cmidrule{1-3}
\multirow{2}{*}{FlexPrefill}
    & $\gamma$=0.9 & 1.70$\times$ \\
    & $\gamma$=0.99 & 1.14$\times$ \\
\cmidrule{1-3}
\multirow{2}{*}{BLASST$^{\star}$}
    & 50\% & 1.14$\times$ \\
    & 70\% & 1.17$\times$ \\
\bottomrule
\end{tabular}
\end{table}

\subsection{Extra Hardware Setup Details}
\label{app:hardware}

\paragraph{A6000.} The A6000 (GA102, SM~8.6) provides $\sim\!768$~GB/s of memory bandwidth versus $\sim\!2{,}039$~GB/s on the A100 (GA100, SM~8.0), with FP16 tensor-core throughput also lower (76 vs.\ 312 TFLOPS). To run on SM~8.6, the kernel was rebuilt with \texttt{FLASH\_ATTN\_CUDA\_ARCHS="8.6"}. Crucially, we reuse the threshold tensor calibrated on A100 with no recalibration. Speedups in main Table~\ref{tab:extra-hardware} are measured on RULER 64K with 50 samples.

\paragraph{H100.} We port BSFA to H100 (Hopper, SM~90) using block dimensions $B_M=128$, $B_N=224$ to match Hopper's memory hierarchy, with the threshold tensor recalibrated for the new block size. We have implemented BSFA on H100 on top of the same attention algorithm used by FA2. BSFA's porting into FA-3 is left to future work. Speedups in main Table~\ref{tab:extra-hardware} are measured on LongBench longest-10 narrativeqa samples ($\sim$65K tokens each).

\subsection{Qwen2.5-7B Details}
\label{app:qwen}

We evaluate on Qwen2.5-7B-Instruct~\citep{qwen2025qwen25technicalreport} with 28 layers, 28 attention heads, 4 KV heads (GQA-7), and head dimension 128. The threshold tensor stores three sparsity levels ($k\in\{16, 32, 64\}$) and totals 2.3~MB. Main Table~\ref{tab:qwen} reports the resulting LongBench accuracy and RULER 64K speedup.

\subsection{Baseline Implementation Details}
\label{app:baselines}

We document method-specific implementation choices for the sparse-attention baselines.

\paragraph{SpargeAttention.} We use the \texttt{spas\_sage\_attn\_meansim\_topk\_cuda} A100 kernel, which combines sparsity with INT8 Q/K quantization. The quantization is orthogonal to our FP16 selective computation. We reached out to the authors to verify our results but received no response.

\paragraph{XAttention.} We use the hardcoded LLaMA-3.1-8B thresholds in their \texttt{eval/LongBench/ratio.py}. The DP optimizer described in their~\S2.3 is not in the released code (their GitHub issue~\#19).

\paragraph{FlexPrefill.} FlexPrefill requires bf16, so its TTFT is reported against a dtype-matched bf16 dense baseline.

\section{Background Details}
\label{app:background}

\subsection{Scaled Dot-Product Attention}
\label{app:attention}

Transformers use multi-head scaled dot-product attention to process sequences of tokens \citep{vaswani2023attentionneed}. Given a sequence of $N$ tokens with model dimension $d_{\text{model}}$, the input $\mathbf{X} \in \mathbb{R}^{N \times d_{\text{model}}}$ is projected into $H$ attention heads, each with dimension $d = d_{\text{model}}/H$. For each head $h$:
\begin{align}
\mathbf{Q}^{(h)} &= \mathbf{X}\mathbf{W}_Q^{(h)} \in \mathbb{R}^{N \times d} \quad \text{(queries)} \\
\mathbf{K}^{(h)} &= \mathbf{X}\mathbf{W}_K^{(h)} \in \mathbb{R}^{N \times d} \quad \text{(keys)} \\
\mathbf{V}^{(h)} &= \mathbf{X}\mathbf{W}_V^{(h)} \in \mathbb{R}^{N \times d} \quad \text{(values)}
\end{align}
where $\mathbf{W}_Q^{(h)}, \mathbf{W}_K^{(h)}, \mathbf{W}_V^{(h)} \in \mathbb{R}^{d_{\text{model}} \times d}$ are learned projection matrices. The attention output for each head is:
\begin{align}
\nonumber\text{Attention}^{(h)}(\mathbf{Q}^{(h)}, \mathbf{K}^{(h)}, \mathbf{V}^{(h)}) = \\\underbrace{\mathrm{softmax}\left(\frac{\mathbf{Q}^{(h)}(\mathbf{K}^{(h)})^\top}{\sqrt{d}}\right)}_{\mathbf{P}^{(h)} \in \mathbb{R}^{N \times N}} \mathbf{V}^{(h)}
\end{align}

The outputs from all heads are concatenated (yielding dimension $d_{\text{model}} = H \cdot d$) and then passed through a final linear projection $\mathbf{W}_O \in \mathbb{R}^{d_{\text{model}} \times d_{\text{model}}}$ to mix information across heads. For simplicity of notation, we omit positional encodings (e.g., RoPE) and focus on a single head in subsequent discussions, dropping the superscript $(h)$.

In standard implementations, the projection matrices $\mathbf{W}_Q$, $\mathbf{W}_K$, $\mathbf{W}_V$ are applied to the full $d_{\text{model}}$-dimensional input before splitting into heads. The computational costs are:
\begin{itemize}
\item \textbf{Linear projections}: $O(Nd_{\text{model}}^2)$ FLOPs total for all $\mathbf{Q}$, $\mathbf{K}$, $\mathbf{V}$ projections across all heads.
\item \textbf{Score computation (QK)}: $O(N^2d)$ FLOPs per head, $O(N^2d_{\text{model}})$ total.
\item \textbf{Softmax normalization}: $O(N^2)$ operations per head, $O(N^2H)$ total.
\item \textbf{Value aggregation (PV)}: $O(N^2d)$ FLOPs per head, $O(N^2d_{\text{model}})$ total.
\item \textbf{Output projection}: $O(Nd_{\text{model}}^2)$ FLOPs.
\end{itemize}

For long sequences where $N \gg d_{\text{model}}$, the operations quadratic in $N$ dominate: both the QK score computation and PV aggregation scale as $O(N^2d_{\text{model}})$ while the linear projections scale as $O(Nd_{\text{model}}^2)$. For example, in Llama-3.1-8B with $d_{\text{model}} = 4096$ ($d = 128$, $H = 32$), processing a sequence of $N = 128$K tokens requires $N^2d_{\text{model}} \approx 6.7 \times 10^{13}$ operations for each of QK and PV, while the linear projections require only $Nd_{\text{model}}^2 \approx 2.1 \times 10^{12}$ operations, yielding a ratio of approximately 32:1. Therefore, optimizing attention for long contexts primarily requires addressing the $O(N^2d)$ complexity of the QK and PV operations.

Beyond computational complexity, memory becomes an equally critical bottleneck. Materializing the full $N \times N$ attention matrix $\mathbf{P}$ requires $O(N^2)$ memory, which becomes prohibitive for long sequences. For instance, storing the attention scores for our 128K example would require over 16 billion elements per head, consuming 32GB in FP16 precision for a single head alone.

\subsection{FlashAttention-2: Tiled Computation}
\label{app:flashattention}

FlashAttention-2 \citep{dao2023flashattention2fasterattentionbetter} computes exact attention without materializing the full $N \times N$ score matrix by using a tiled algorithm with online softmax. The key insight is to partition the computation into blocks and maintain running statistics that allow incremental updates. The algorithm leverages the GPU memory hierarchy, keeping intermediate results in fast on-chip SRAM (Static RAM) while streaming data from slower but larger HBM (High Bandwidth Memory).

The algorithm partitions the query sequence into $M_Q = \lceil N/B_M \rceil$ blocks of size $B_M$ and the key/value sequences into $M_{KV} = \lceil N/B_N \rceil$ blocks of size $B_N$. Each query block $\mathbf{Q}_i \in \mathbb{R}^{B_M \times d}$ (for $i = 0, \ldots, M_Q - 1$) processes all key blocks $\mathbf{K}_j \in \mathbb{R}^{B_N \times d}$ and corresponding value blocks $\mathbf{V}_j \in \mathbb{R}^{B_N \times d}$ (for $j = 0, \ldots, M_{KV} - 1$) in sequence.

The key algorithmic innovation is the use of online softmax with incremental updates. Instead of computing and storing the full attention matrix, FlashAttention processes one block at a time and maintains running statistics (maximum values and normalizers) that are updated as each new block is processed. Conceptually, for each query block $\mathbf{Q}_i$, the algorithm: (1) loads $\mathbf{Q}_i$ from HBM to on-chip memory (shared SRAM and registers) where it remains throughout processing, (2) streams key blocks $\mathbf{K}_j$ from HBM one at a time, (3) computes attention scores $\mathbf{S}_{ij} = \mathbf{Q}_i\mathbf{K}_j^\top / \sqrt{d}$ using tensor cores, with score tiles kept on-chip, (4) updates running statistics (maximum values and normalizers) which are also maintained on-chip, (5) loads the corresponding value block $\mathbf{V}_j$ and accumulates the weighted output in on-chip memory, and (6) rescales the accumulated results to maintain numerical precision. Throughout the inner loop over key/value blocks, all intermediate results stay on-chip. Only after processing all blocks for a given query block $\mathbf{Q}_i$ is the final output $\mathbf{O}_i$ written back to HBM. The score tiles $\mathbf{S}_{ij}$ are never written to HBM but exist only in on-chip memory during computation. This tiled approach means the full $N \times N$ attention scores matrix is never materialized in memory, fundamentally changing the memory requirements from quadratic to linear. The complete algorithmic details can be found in \citet{dao2023flashattention2fasterattentionbetter}.

The per-tile computational cost reveals why attention remains expensive even with FlashAttention's memory optimizations. For each tile $(i,j)$, the algorithm performs two equally expensive matrix multiplications: computing the attention scores $\mathbf{S}_{ij} = \mathbf{Q}_i\mathbf{K}_j^\top$ and applying them to values $\tilde{\mathbf{P}}_{ij}\mathbf{V}_j$, each requiring $2B_M B_N d$ FLOPs. The algorithm must also load both $\mathbf{K}_j$ and $\mathbf{V}_j$ blocks from HBM, with each block containing $B_N d$ elements. Additional overhead from computing exponentials and maintaining running statistics is comparatively minor ($O(B_M B_N)$ operations). Since FlashAttention processes $M_Q \times M_{KV}$ total tiles (approximately half for causal attention), and the QK and PV operations equally dominate the computational cost, each contributes roughly $2N^2d$ FLOPs to the total complexity.

\subsection{Calibration and Practical Details}
\label{app:calibration}

\paragraph{Threshold tensor.} We calibrate a tensor $\mathbf{T} \in \mathbb{R}^{S \times L \times H \times \lceil N_{\max}/B_M \rceil}$ where $S$ is the number of sparsity levels, $L$ is the number of layers, $H$ is the number of heads, and $N_{\max}$ is the maximum calibration sequence length. At inference, selecting sparsity level $k$ extracts the slice $\mathbf{T}[k, :, :, :]$, which provides per-layer, per-head, per-position thresholds.

\paragraph{Storage overhead.} Storage is minimal: for a single sparsity level and $N_{\max}=65{,}536$, Llama-3.1-8B requires approximately $5.2\times10^5$ threshold values, negligible compared to the model's 8B parameters ($0.007\%$).

\paragraph{Practical considerations.} For sequences not aligned with block boundaries, we pad by repeating the last token to avoid introducing artificial scores that would distort the top-$k$ selection. Sequences exceeding $N_{\max}$ reuse thresholds from position $N_{\max}$, as attention patterns empirically stabilize in later positions. While presented for causal attention, BSFA extends naturally to bidirectional settings.

\paragraph{Deployment flexibility.} Storing multiple sparsity levels enables dynamic adjustment of the latency-accuracy trade-off at deployment time without recalibration. Different applications may prioritize speed versus accuracy differently, and specific tasks like information retrieval can use much higher sparsity while maintaining accuracy.

\end{document}

%% file: math_commands.tex
\usepackage{amsmath,amsfonts,bm}
\usepackage{nicefrac}

\def\eqref#1{equation~\ref{#1}}

\def\1{\bm{1}}

\DeclareMathAlphabet{\mathsfit}{\encodingdefault}{\sfdefault}{m}{sl}
\SetMathAlphabet{\mathsfit}{bold}{\encodingdefault}{\sfdefault}{bx}{n}

\usepackage{xspace}
\makeatletter
\DeclareRobustCommand\onedot{\futurelet\@let@token\@onedot}
\def\@onedot{\ifx\@let@token.\else.\null\fi\xspace}

\makeatother